\documentclass[lettersize,journal]{IEEEtran}
\usepackage{amsmath,amsfonts}
\usepackage{algorithmic}
\usepackage{algorithm}
\usepackage{array}
\usepackage[caption=false,font=normalsize,labelfont=sf,textfont=sf]{subfig}
\usepackage{textcomp}
\usepackage{stfloats}
\usepackage{url}
\usepackage{verbatim}
\usepackage{graphicx}
\usepackage{cite}
\usepackage{colortbl}
\usepackage{xcolor}
\usepackage{multirow}
\usepackage{multicol}
\usepackage{hyperref}
\begin{document}

\title{I Speak and You Find: Robust 3D Visual Grounding with Noisy and Ambiguous Speech Inputs}

\author{Yu Qi, Lipeng Gu, Honghua Chen, Liangliang Nan, and Mingqiang Wei,~\textit{Senior Member, IEEE}
\thanks{Y. Qi, L. Gu, H. Chen and M. Wei are with the School
of Computer Science and Technology, Nanjing University of Aeronautics
and Astronautics, Nanjing, China (e-mail: qiyuyu@nuaa.edu.cn; glp1224@163.com; chenhonghuacn@gmail.com; mingqiang.wei@gmail.com).}
\thanks{L. Nan is with the Urban Data Science Section, Delft University of Technology, Delft, Netherlands (e-mail: liangliang.nan@tudelft.nl).}
}

\markboth{Journal of \LaTeX\ Class Files,~Vol.~14, No.~8, August~2021}%
{Shell \MakeLowercase{\textit{et al.}}: A Sample Article Using IEEEtran.cls for IEEE Journals}


\maketitle

\begin{abstract}
Existing 3D visual grounding methods rely on precise text prompts to locate objects within 3D scenes. Speech, as a natural and intuitive modality, offers a promising alternative. Real-world speech inputs, however, often suffer from transcription errors due to accents, background noise, and varying speech rates, limiting the applicability of existing 3DVG methods. 
To address these challenges, we propose \textbf{SpeechRefer}, a novel 3DVG framework designed to enhance performance in the presence of noisy and ambiguous speech-to-text transcriptions. SpeechRefer integrates seamlessly with existing 3DVG models and introduces two key innovations. 
First, the Speech Complementary Module captures acoustic similarities between phonetically related words and highlights subtle distinctions, generating complementary proposal scores from the speech signal. This reduces dependence on potentially erroneous transcriptions. Second, the Contrastive Complementary Module employs contrastive learning to align erroneous text features with corresponding speech features, ensuring robust performance even when transcription errors dominate.
Extensive experiments on the SpeechRefer and SpeechNr3D datasets demonstrate that SpeechRefer improves the performance of existing 3DVG methods by a large margin, which highlights SpeechRefer's potential to bridge the gap between noisy speech inputs and reliable 3DVG, enabling more intuitive and practical multimodal systems.
\end{abstract}


\section{Introduction}
\label{sec:introduction}

\IEEEPARstart{I}{nteracting} with visual data using natural language is a rapidly advancing field\cite{visual-program,denseAV,vision-language-navigation-HAMT,lu2024scaneru,li2021bridging,chen2023dialogmcf,parekh2019weakly} with real-world applications such as robotics and AR/VR. Within this domain, 3D visual grounding, commonly referred to as text-guided 3D visual grounding (T-3DVG), focuses on locating specific objects based on precise text prompts within 3D scenes.

Recent research\cite{d3net,G3LQ,MA2TransVG,butd-detr,3djcg,Shi_2024_CVPR,chen2023unit3d,guo2023viewrefer,huang2022multi} has focused on designing sophisticated architectures to achieve state-of-the-art performance in 3DVG. For example, \cite{G3LQ} introduces a framework that explicitly models geometry-aware visual representations while generating fine-grained, language-guided object queries. \cite{MA2TransVG} learns the multi-attribute interactions to refine the intra-modal and inter-modal grounding cues. Although accurate text features offer structured word-level context for strong performance in T-3DVG tasks, they often fail to address challenges in complex real-world environments and overlook the actuality that precise text cannot appear out of thin air.
\begin{figure}
\centering
\includegraphics[width=0.5\textwidth]{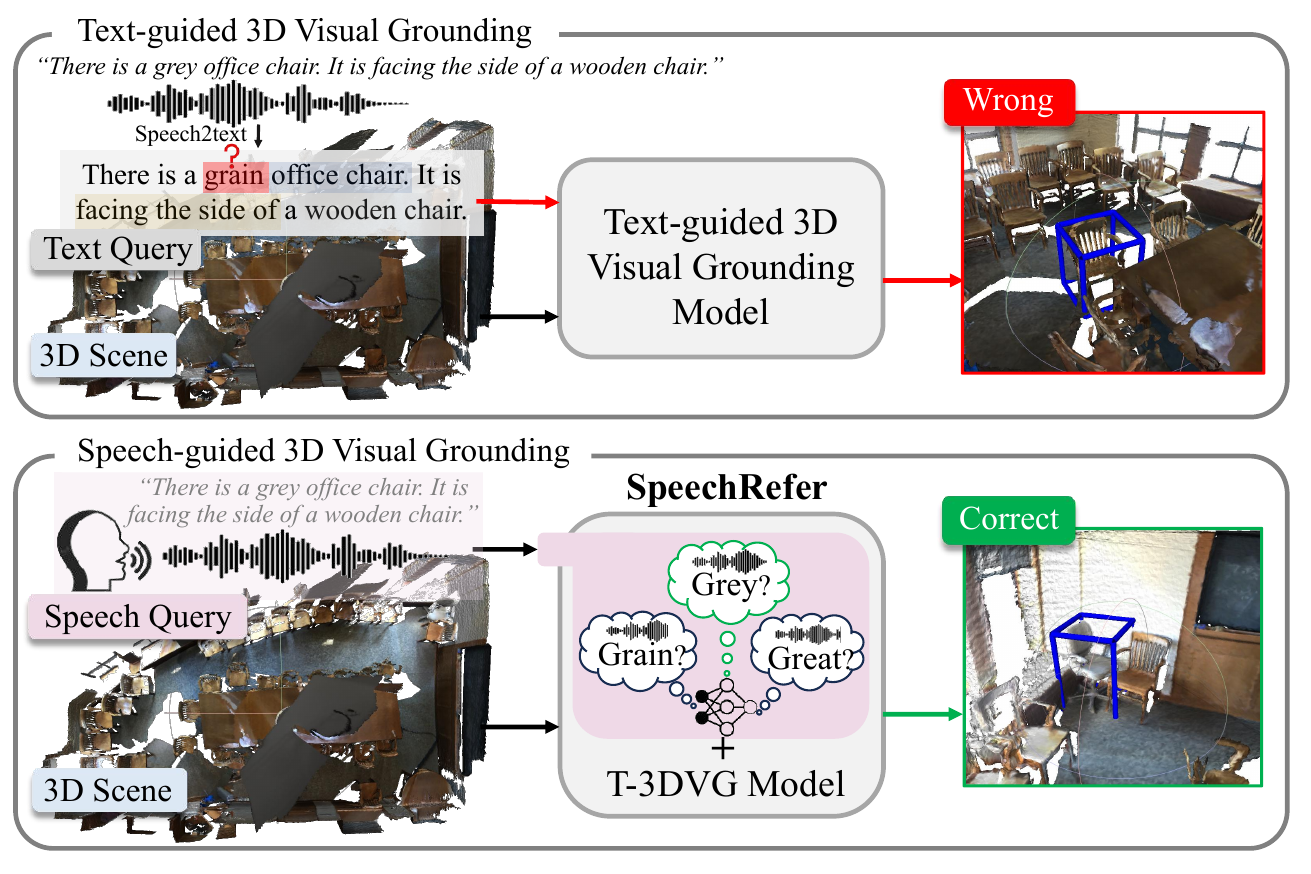}
\caption{\textbf{Illustration of SpeechRefer.} SpeechRefer captures acoustic similarities between phonetically related words, which acts as a complement to potential transcription errors, reducing dependence on potentially erroneous transcriptions.}
\label{fig:introduction}
\end{figure}

Considering the real-world environments, speech stands out as a natural modality compared to text, offering an intuitive way for humans to interact with AI systems. For example, you can directly ask a robot to locate ``\textit{the brown guitar next to the bed}", which reflects natural human communication in real-world scenarios. 
Logically, speech recognition models can serve as a bridge between speech and existing T-3DVG methods to achieve speech-guided 3D visual grounding (S-3DVG). Despite advances in speech recognition, complex real-world environments inevitably lead to a significant challenge during transcriptions: errors and uncertainties that arise due to variations in speech quality, such as accents, background noise, or speech rate.
For instance, as illustrated in Figure \ref{fig:introduction}, the input speech with Chinese accent ``\textit{There is a grey office chair}” might be transcribed incorrectly as ``\textit{There is a grain office chair}". This misinterpretation leads the model to search for a non-existent object (attribute) and fails to ground the correct object. Such transcription errors are common and have cascading effects, severely impairing the system's ability to ground objects accurately. 
Unfortunately, existing T-3DVG methods rely heavily on precise text inputs to perform effectively, which reveals a significant gap between T-3DVG methods and real-world scenarios with complexity and noise.


To address this challenge, we propose \textbf{SpeechRefer}, a novel 3DVG framework that seamlessly integrates with existing T-3DVG methods, leveraging the features of raw speech to complement text derived from the speech and mitigate the impact of transcription errors.
As shown in Figure \ref{fig:introduction}, the speech features space preserves acoustic similarities between phonetically similar words (e.g. \textit{grey, grain}) while capturing phonetic nuances often lost during speech-to-text conversion. This allows the system to infer the correct target object by considering alternative interpretations of the user's intent, rather than relying solely on potentially erroneous transcription. 

Specifically, we propose a speech complementary module comprising two components: a phonetic-aware refinement module and a confidence-based complementary module. The phonetic-aware refinement module inherently captures acoustic similarities between phonetically related words while preserving subtle speech nuances. Then, the confidence-based complementary module explicitly generates complementary proposal scores based on speech features, effectively complementing the transcribed text scores and reducing reliance on potentially erroneous transcription. By incorporating complementary contextual information, the system identifies ``\textit{grey}" as the most likely intended attribute, even when the transcription incorrectly suggests ``\textit{grain}". 
To address scenarios where erroneous text dominates and renders speech complementary scores ineffective, we further propose a contrastive complementary module. This module uses contrastive learning to align erroneous text features with accurate speech features, ensuring that speech features effectively complement text derived from the speech even in challenging situations. 
By integrating speech features as a complement to textual features, SpeechRefer effectively addresses the challenges of imperfect speech-to-text transcriptions, which enables more accurate interpretation of user intent in uncertain conditions

Our contributions include:
\begin{itemize}
    \item SpeechRefer is the first and novel 3DVG framework designed to enhance model robustness in the presence of noisy and ambiguous transcriptions, which can integrate with existing T-3DVG methods and improve its performance by a large margin.
    \item The speech complementary module retains acoustic similarities while distinguishing subtle differences, avoiding being constrained by potential transcription errors. The contrastive complementary module corrects misaligned text features using speech features, ensuring robust performance even in transcription-dominated scenarios.
    \item We introduce the first speech datasets that reflect complex real-world environments, representing a promising step toward more intuitive 3D visual grounding.
\end{itemize}
\begin{figure*}
\centering
\includegraphics[width=\textwidth]{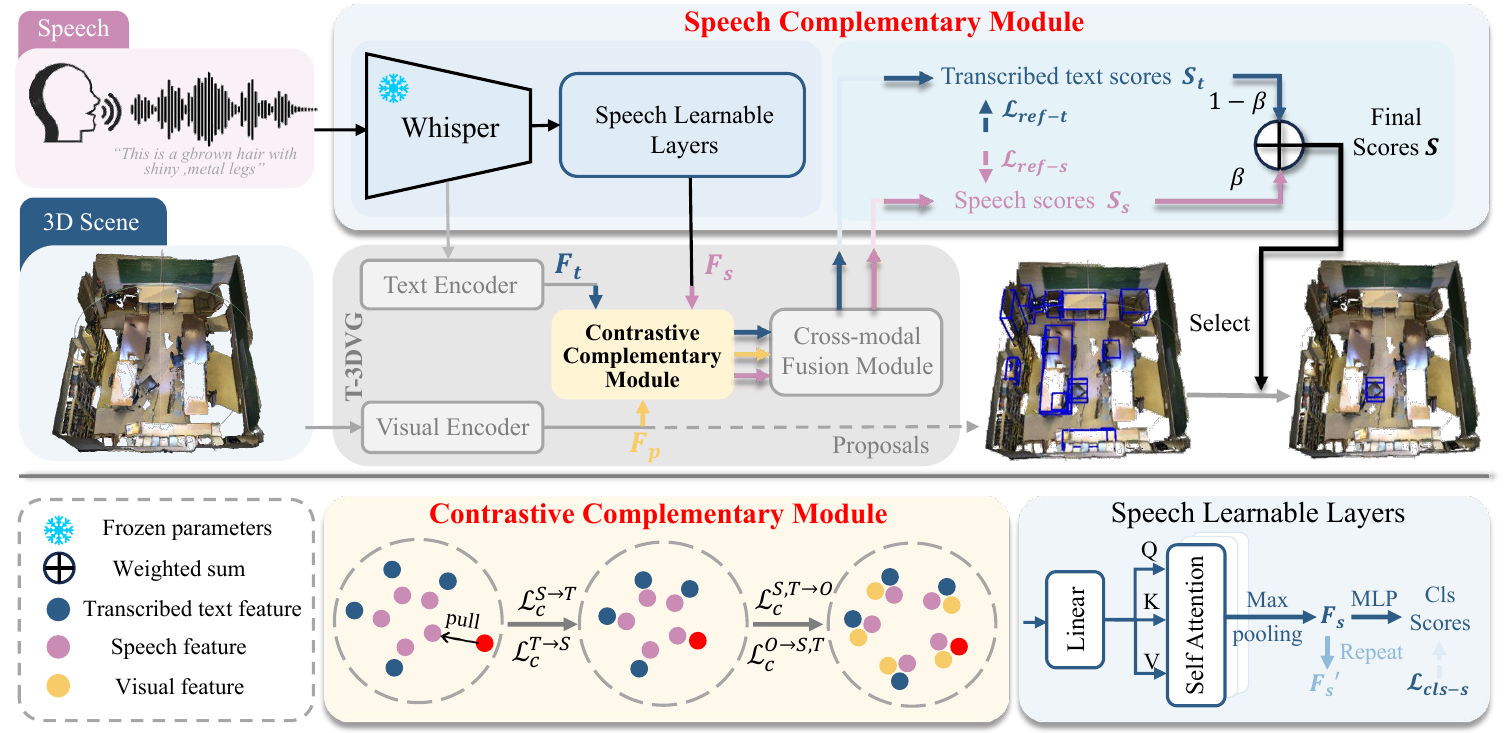}
\caption{\textbf{Overview of SpeechRefer.} The framework builds on existing T-3DVG methods, incorporating two novel modules highlighted here: (1) \textbf{Speech complementary module}: This module encodes speech features $F_s$ and generates speech scores $S_s$ to explicitly complement the transcribed text. By integrating speech features, it helps identify the correct object and mitigates the impact of potential transcription errors. (2) \textbf{Contrastive complementary module}: This module aligns potentially erroneous text features with corresponding speech features by contrastive learning. This alignment ensures robust performance even when erroneous text features dominate. 
Key components are color-coded, where gray represents pre-existing modules (e.g., visual and text encoders). $S$, $T$, and $O$ denote the global speech feature, sentence-level transcribed text feature, and the mean of object features of all target objects associated with the same description, respectively.}
\label{fig:overview}
\end{figure*}
\section{Related Work}

\subsection{Speech Recognition} 
Speech recognition is a technology that converts speech into text and is one of the most critical advancements in AI-driven interaction. Its applications have permeated various aspects of daily life and production, such as voice assistants in mobile phones. After years of research and development, speech recognition has become quite advanced \cite{whisper, whisperX}, particularly for English. For instance, the Whisper\cite{whisper} model achieves a word error rate (WER) as low as 4.1\% on the English category of the Fleurs dataset. However, even with state-of-the-art speech recognition models, transcription errors are still common due to variations in speech quality, such as accents, background noise, or speech rate. Potential misinterpretation leads the model to search for a non-existent object (attribute) and fails to ground the correct object. SpeechRefer was proposed to address this challenge.

\subsection{Language and Speech in 3D Interaction}
Recent works in 3D vision-language understanding have advanced tasks like object grounding, dense captioning, and question answering in 3D scenes. Mao et al.\cite{mao2023complete} proposed a modality alignment network for 3D dense captioning, capturing both local and global spatial relationships to improve semantic alignment. Ye et al.\cite{ye20223d} introduced 3DQA, extending traditional VQA into 3D environments by jointly modeling geometry, appearance, and language inputs
. These efforts highlight the potential of language as a powerful interface for 3D scene understanding.

Meanwhile, audio-visual learning has been increasingly explored in 3D tasks such as 3D facial animation\cite{li2024pose} and embodied interaction, where speech serves as a driving signal for motion generation or behavior control. This line of work highlights the growing interest in leveraging speech as a natural modality for enhancing realism, interactivity, and user engagement in 3D environments.

Inspired by these directions, our work explores speech-guided 3D visual grounding under noisy conditions—a novel yet promising task. By enhancing robustness to speech ambiguities, we aim to support more intuitive, speech-based 3D interaction in real-world applications such as AR/VR.

\subsection{3D Visual Grounding} 
Since the introduction of the T-3DVG task by ScanRefer\cite{scanrefer} and ReferIt3D\cite{referit3d}, it has developed rapidly and found applications across various fields. Existing T-3DVG methods are typically categorized into two-stage and one-stage frameworks, with the majority adopting a two-stage framework. Two-stage methods\cite{liu2021refer,yuan2021instancerefer,sat,tgnn,3D-vista-scene-understanding,3D-VLP2023CVPR} first extract text features from the query language using language models\cite{pennington2014glove, chung2014empirical} and generate proposal boxes using a pre-trained detector\cite{liu2021group, houghvote}. These features are then fused to select the best-matched object. Among these, ConcreNet\cite{ConcreteNet} presents four novel stand-alone modules designed to enhance performance in challenging repetitive instances. MA2TransVG\cite{MA2TransVG} learns the multi-attribute interactions to refine the intra-modal and inter-modal grounding cues. Conversely, one-stage methods\cite{3d-sps,EDA,3DVLP2024AAAI} directly localize the target in a single step. Notably, 3D-SPS\cite{3d-sps}, the first one-stage model, utilizes text features to guide visual keypoint selection, facilitating progressive object grounding. EDA\cite{EDA} introduces a text decoupling module to generate textual features for each semantic component, significantly enhancing grounding performance, and even surpassing most existing two-stage frameworks.

In general, most methods focus on designing sophisticated architectures to achieve state-of-the-art grounding performance, but they overlook the challenges posed by complex real-world environments and practical applications. Our SpeechRefer is the first and novel 3DVG framework that could integrate with existing T-3DVG methods, bridging the gap between T-3DVG methods and real-world scenarios with complexity and noise.

\section{Method}
\label{sec:method}
\subsection{Overview}
As shown in Figure \ref{fig:overview}, our network has two inputs: one is point cloud $P\in \mathbb{R}^{N\times (3+K)}$ that represents the whole 3D scene by 3D coordinates and K-dimensional auxiliary feature (e.g., RGB, normal vectors, or the pre-trained multi-view features\cite{scanrefer}). Another input is speech, which is a human voice describing a target object in the 3D scene. The output is the predicted target object, which is an axis-aligned bounding box with the center $\textbf{c}=[c_x,c_y,c_z]\in\mathbb{R}^3$ in the world coordinate, and the size $\textbf{s} = [s_x,s_y,s_z]\in\mathbb{R}^3$. Our SpeechRefer is a novel S-3DVG framework that could seamlessly integrate with existing T-3DVG methods, so the visual encoder, language encoder, and cross-modal fusion module are consistent with the T-3DVG models. 

Our SpeechRefer features two key innovations. First, the \textbf{speech complementary module} (\ref{Speech Complementary Module}) captures acoustic similarities between phonetically related words while distinguishing subtle distinctions. This module generates complementary proposal scores from speech signals, reducing dependence on potentially erroneous transcriptions. \textbf{Contrastive complementary module} (\ref{Contrastive Complementary Module}) aligns
erroneous text features with corresponding correct speech
features through contrastive learning, ensuring robust performance even when errors dominate. Finally, Section \ref{finalloss} describes the total loss. 
\subsection{Baseline: Text-guided 3D Visual Grounding via Speech-to-Text} 
\label{sec:grounding with Transcribed Text}
\begin{figure*}
\centering
\includegraphics[width=\textwidth]{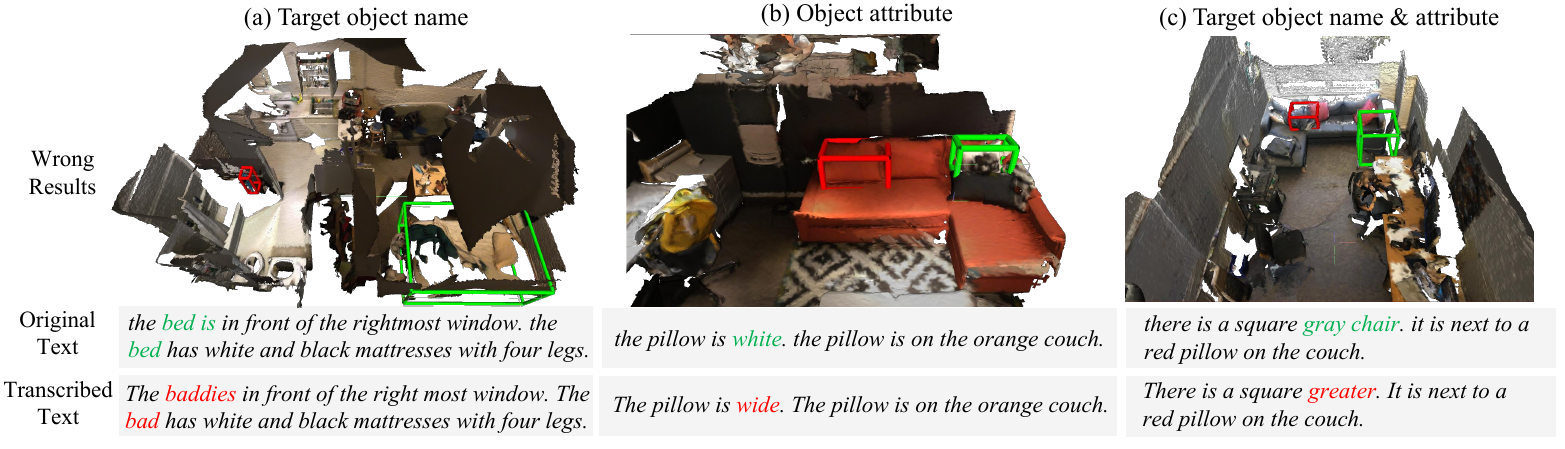}
\caption{Incorrectly transcribed text may include errors such as misinterpreting the target object name or its attributes. These mistakes can mislead the model into searching for a non-existent object or attribute, ultimately failing to correctly identify and ground the intended object.}
\label{fig:errors}
\end{figure*}

To clarify the challenges addressed by our approach, we first describe the baseline T-3DVG method that relies on speech-to-text transcription. The speech recognition model serves as a bridge between speech and T-3DVG by converting speech to text, which is then processed by any T-3DVG model to locate objects. However, even advanced speech recognition models frequently introduce critical errors due to factors like accents, background noise, and speech rate, reflecting real-world conditions. Figure \ref{fig:errors} illustrates common transcription errors when the input speech contains uncertainties such as accents, where the original T-3DVG model often fails to ground the correct objects. Among these errors, inaccuracies in transcribing target object names are particularly frequent and problematic. Prior studies\cite{EDA} have confirmed that grounding results significantly degrade when object names are omitted. Additionally, errors in attributes, such as misinterpreting ``\textit{white}" as ``\textit{wide}" can also mislead the model by overlooking crucial contextual cues, increasing its susceptibility to distractors within the same category and ultimately impairing overall performance. 

\subsection{Speech Complementary Module}
\label{Speech Complementary Module}
To address transcription errors, we propose a speech complementary module that incorporates phonetic features directly from speech. It draws on the acoustic similarities and nuances of input speech as a complement to mitigate the negative effects of potentially erroneous transcriptions. It mainly comprises two submodules.
\begin{figure}
\centering
\includegraphics[width=0.45\textwidth]{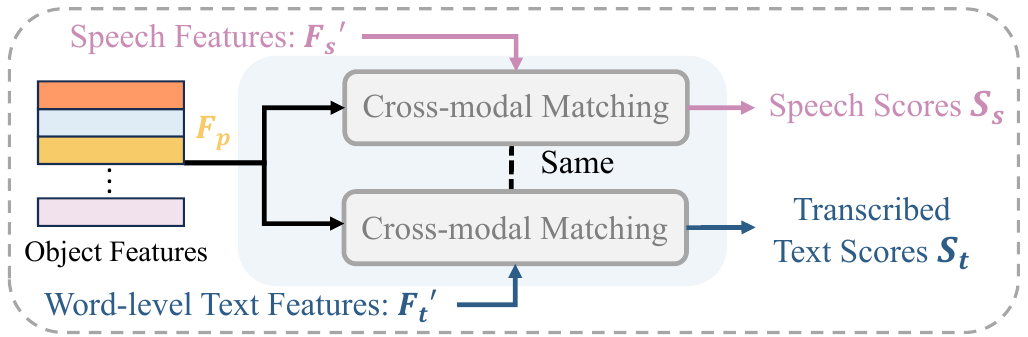}
\caption{Cross-modal Fusion Module. The module consists of two identical cross-modal matching modules, which is the specific matching module of existing T-3DVG models.}
\label{fig:fusion module}
\end{figure}\\
\textbf{Phonetic-aware Refinement Module.}
\label{PRM} 
First, we utilize the Whisper\cite{whisper} as our speech encoder, which inherently preserves information about acoustic similarities that are often lost during the speech-to-text conversion. Whisper is a general-purpose speech recognition model trained on a large, diverse speech dataset. As a multitask model, it can also perform language identification and speech translation, effectively exploiting contextual clues in speech, which makes it an excellent source of initial speech features for a wide range of downstream networks. Before the raw speech is fed into the network, it's resampled to 16,000 Hz, and an 80-channel log-magnitude Mel spectrogram representation $M_s\in \mathbb{R}^{80 \times L}$ is computed on 25 millisecond windows with a stride of 10 milliseconds\cite{whisper}. The speech log-magnitude Mel spectrogram $M_s$ is encoded into speech feature $W_s\in \mathbb{R}^{N_s\times D_s}$ through the Whisper encoder, where $N_s$ denotes the temporal sequence length of speech, and $D_s$ denotes semantic feature dimension.

To enhance the pre-trained Whisper model, we design learnable layers and introduce classification loss to fine-tune the speech features. This process emphasizes the speech features of the target object class, which enables the model to better distinguish subtle differences. The initial speech feature $W_s$ is the input of learnable layers, which can be formulated as: 
\begin{equation}
    W_s^{\prime} = SelfAttention(Linear(W_s))
\end{equation}
where Linear(·) denotes the linear projection, SelfAttention(·) denotes the multi-head self-attention mechanism\cite{vaswani2017attention}. Then, the global speech feature is extracted via max pooling, denoted as $F_s\in\mathbb{R}^{1\times D_s}$, and the features are repeated and stacked, generating the final speech feature representation $F_s^\prime$.
After that, a speech classification loss is included, which can be formulated as:
\begin{equation}
    \mathcal{L}_{cls-s} = -\sum_{i = 1}^{C}y_i log(\hat{y}_i)
\end{equation}
where C is the number of classes, $y_i$ is the ground truth one-hot encoding, $\hat{y}_i$ is classification scores after a classifier head. 
This enables the model to dynamically capture acoustic similarities in speech while distinguishing subtle distinctions by emphasizing the features of the target object name, further improving the expressiveness of speech features.\\
\textbf{Confidence-based Complementary Module.} 
Based on the former, the confidence-based complementary module explicitly generates complementary scores for proposals based on speech features to complement the textual score, mitigating the impact of potential transcription errors. Figure \ref{fig:fusion module} illustrates the structure, the cross-modal matching component is the specific matching module used in existing T-3DVG models. We utilize the same network to fuse speech-visual and transcribed text-visual features respectively, ensuring feature consistency. Then the speech-visual fusion features are fed into an FFN layer to generate the speech scores $S_s$ for the M generated proposals except for transcribed text scores $S_t$. 
This allows the model to incorporate multiple potential interpretations of the user's intent, leveraging speech scores to complement the transcribed text ones. This reduces the model's reliance on potentially erroneous transcriptions. The final scores $S = \beta S_s + (1 - \beta) S_t$, where $\beta$ is a weighting factor. Eventually, the bounding box with the highest final score will be considered as the final grounding result. Additionally, we introduce a reference loss to supervise the module, encouraging that the proposal with the highest score is closer to the ground truth, thereby promoting more accurate target object selection:
\begin{equation}
    L_{ref}=\alpha_1 L_{ref-s}+\alpha_2 L_{ref-t}
\end{equation}
\begin{equation}
    L_{ref-s} = -\sum_{i = 1}^{M}t_ilog({s_s}_i), L_{ref-t} = -\sum_{i = 1}^{M}t_ilog({s_t}_i) 
\end{equation}
where $L_{ref-s}$ and $L_{ref-t}$ are cross-entropy losses based on speech and transcribed text scores respectively, following the strategy in ScanRefer\cite{scanrefer}, we set the label $t_i$ for the $i^{th}$ box that has the highest IoU score with the ground truth as 1 and others as 0. 


\subsection{Contrastive Complementary Module.}
\label{Contrastive Complementary Module}
While utilizing speech information significantly mitigates the negative impact of potential transcription errors, situations still arise where erroneous text weighs heavily on proposal selection, leading to failures in identifying the target object despite the complementary speech scores. To address this, we propose a contrastive complementary module. 
This module builds on the observation that most objects (or attributes) incorrectly generated by transcription, such as ``\textit{bat}”, ``\textit{grain}”, etc., do not exist in the input scene. At the feature level, even if the textual features represent ``\textit{grain}”, they should align with the speech features of ``\textit{grey}" to correctly infer the target object. To achieve this, we employ contrastive learning to align incorrect textual features with the correct speech features, ensuring more accurate target object inference:
\begin{equation}
    \mathcal{L}_{c}^{T\rightarrow S} = -\sum_{i=1}^{N}log\frac{exp(s(T_i, S_i)/t)}{\sum_{j=1}^{n}exp(s(T_i,S_j)/t)}
\end{equation}
where $N$ is the number of samples in one batch, $t$ is a temperature parameter, and $s(\cdot)$ is the cosine similarity. $S$ is the global speech feature (i.e. $F_s$), and $T$ is the sentence-level transcribed text feature (i.e. $F_t$).

Additionally, to align with the corresponding visual features, we introduce additional contrastive loss between language and visual features: 
\begin{equation}
\begin{aligned}
    \mathcal{L}_{c}^{S,T\rightarrow O} = -( \sum_{i=1}^{N}log\frac{exp(s(S_i, O_i)/t)}{\sum_{j=1}^{n}exp(s(S_i,O_j)/t)} + \\
    \sum_{i=1}^{N}log\frac{exp(s(T_i, O_i)/t)}{\sum_{j=1}^{n}exp(s(T_i,O_j)/t)} )
\end{aligned}
\end{equation}
$O$ is the mean of object features of all target objects paired with the same description. We symmetrize these losses by adding the analogous terms, $\mathcal{L}_{c}^{S\rightarrow T}$, $\mathcal{L}_{c}^{O\rightarrow S, T}$. The total contrastive loss is the average of these contrastive losses. By aligning text features with corresponding speech features, the model can more effectively correct transcription errors, ensuring robust performance even when errors dominate.

\subsection{Total Loss}
\label{finalloss}
We train the network end-to-end with the total loss $\mathcal{L} = \mathcal{L}_{det}+\gamma_1\mathcal{L}_{c} + \gamma_2\mathcal{L}_{ref} + \gamma_3\mathcal{L}_{cls}$. $\mathcal{L}_{det}$ varies with different integration methods. For example, in 3DVG-Transformer, $\mathcal{L}_{det} = 10\mathcal{L}_{vote-reg}+\mathcal{L}_{objn-cls}+\mathcal{L}_{sem-cls}+10\mathcal{L}_{box}$, and this object detection loss exactly follows the loss used in \cite{houghvote} for the ScanNet dataset\cite{ScanNet}. To better adapt the total loss function to different methods, we adjust $\gamma_1, \gamma_2, \gamma_3$ to ensure that all the loss components are balanced and remain in the same order of magnitude.

\begin{table*}
        \renewcommand\arraystretch{1}
	\centering
        \caption{Results on the SpeechRefer\_CA dataset (i.e., SpeechRefer dataset with \textbf{C}hinese \textbf{A}ccent) and comparison results against baseline. Notably, SpeechRefer exhibits promising performance when fine-tuned on the SpeechRefer\_VF dataset (\textbf{V}ariations-\textbf{f}ree SpeechRefer dataset).}
	{\begin{tabular}{ccccccccc}	
            \noalign{\smallskip}\hline\noalign{\smallskip}
            \rowcolor{gray!20}
             & & &\multicolumn{2}{c}{Unique(19\%)} & \multicolumn{2}{c}{Multiple(81\%)} & \multicolumn{2}{c}{Overall}\\
            \rowcolor{gray!20}
		 \multirow{-2}*{Methods} &\multirow{-2}*{Setting} &\multirow{-2}*{Modality} & 0.25 & 0.5 & 0.25 & 0.5 & 0.25 & 0.5 \\
            \noalign{\smallskip}\hline\noalign{\smallskip}
            ScanRefer & Baseline & 3D & 67.93 & 46.87 & 30.99 & 21.39 & 38.15 & 26.34 \\
            3D-VisTA & Baseline & 3D & 79.19 & 72.42 & 42.63 & 38.34 & 49.40 & 44.64 \\             \noalign{\smallskip}\hline\noalign{\smallskip}
            \multirow{5}{*}{3DVG-Transformer}& Baseline & \multirow{5}{*}{3D} & 75.93 & 55.21 & 37.27 & 27.02 & 44.77 & 32.49 \\
            &+SpeechRefer (ours) &  & \textbf{77.13} & \textbf{57.19} & \textbf{39.74} & \textbf{28.73} & \textbf{47.00} & \textbf{34.25}\\
            & Improvements & & \textbf{+1.20} & \textbf{+1.98} & \textbf{+2.47} & \textbf{+1.71} & \textbf{+2.23} & \textbf{+1.76} \\
            \noalign{\smallskip}\cline{2-2}\cline{4-9}\noalign{\smallskip}
            & Fine-tune on SpeechRefer\_VF& & 79.02 & 57.58 & 39.88 & 29.14 & 47.47 & 34.66 \\ 
            \noalign{\smallskip}\hline\noalign{\smallskip}
            \multirow{4}{*}{M3DRefCLIP}& Baseline & \multirow{4}{*}{3D+2D} & 84.60 & \textbf{74.50} & 42.00 & 35.20 & 50.30 & 42.80\\
            &+SpeechRefer (ours)& & \textbf{85.20} & 74.10 & \textbf{44.80} & \textbf{37.30} & \textbf{52.60} & \textbf{44.40} \\
            &Improvements& & \textbf{+0.60} & -0.40 & \textbf{+2.80} & \textbf{+2.10} & \textbf{+2.30} & \textbf{+1.60} \\
            \noalign{\smallskip}\cline{2-2}\cline{4-9}\noalign{\smallskip} 
            & Fine-tune on SpeechRefer\_VF& & 85.90 & 75.30 & 46.10 & 38.00 & 53.80 & 45.20\\ 
		\noalign{\smallskip}\hline
	\end{tabular}}
        \label{tab:scanrefer results}
\end{table*}
\begin{table*}
        \renewcommand\arraystretch{1}
	\centering
        \caption{Results on SpeechNr3D\_CA dataset set with GT boxes and comparison results with the baseline.}
	{\begin{tabular}{cccccccc}	
            \noalign{\smallskip}\hline\noalign{\smallskip}
            \rowcolor{gray!20}
             Methods & Setting & Modality & Overall&Easy & Hard & View Dep & View Indep.\\
            \noalign{\smallskip}\hline\noalign{\smallskip}
            \multirow{4}{*}{M3DRefCLIP}& Baseline & \multirow{4}{*}{3D+2D} & 42.64 & 48.78 & 36.98 & 36.26 & 46.07 \\
            &+SpeechRefer (ours)& & \textbf{46.24} & \textbf{52.62 }& \textbf{40.34} & \textbf{39.80} & \textbf{49.69} \\
            &Improvements& &\textbf{+3.60} & \textbf{+3.84} & \textbf{+3.36} & \textbf{+3.54} & \textbf{+3.62} \\
            \noalign{\smallskip}\cline{2-2}\cline{4-8}\noalign{\smallskip}
            & Fine-tune on SpeechNr3D\_VF& & 47.70 & 54.25 & 41.66 & 39.81& 51.73\\ 
		\noalign{\smallskip}\hline
	\end{tabular}}
        \label{tab:nr3d results}
\end{table*}
\section{Experiments}
\label{sec:experiments}
\subsection{Experimental Setting} 
\begin{figure}
    \centering
    \includegraphics[width=\linewidth]{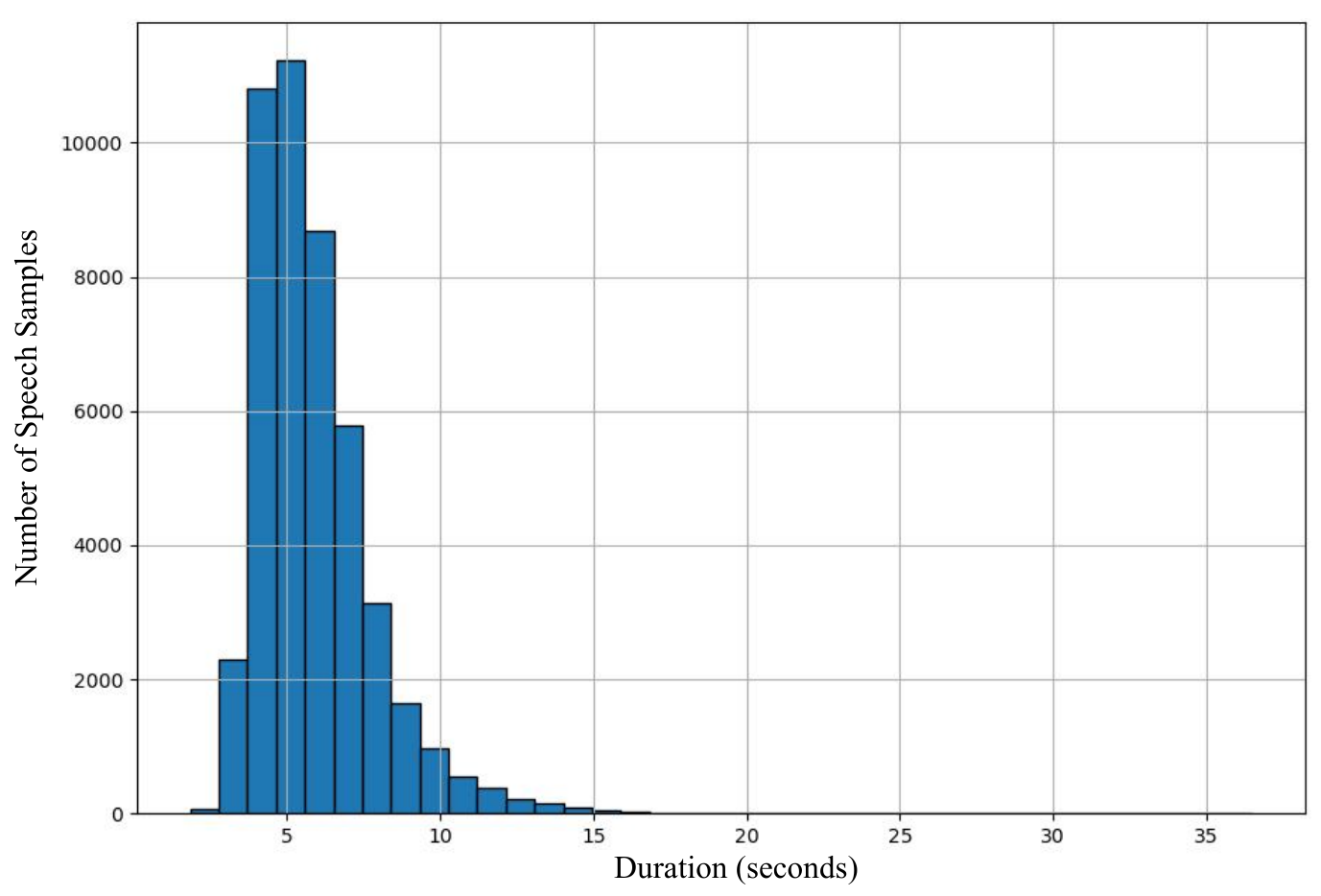}
    \caption{Speech Duration Distribution of SpeechRefer dataset.}
    \label{fig:statistical}
\end{figure}

\textbf{SpeechRefer and SpeechNr3D datasets.} We introduce new speech datasets designed to simulate real-world variations, such as accents, speech rates, and background noise, based on ScanRefer\cite{scanrefer} and Nr3D\cite{referit3d}. To incorporate accents, we use the off-the-shelf text-to-speech tool\footnote{https://pypi.org/project/pyttsx3/} to generate speech with a \textbf{C}hinese \textbf{a}ccent, creating the SpeechRefer\_CA and SpeechNr3D\_CA datasets. For background noise, we overlay one of five common sounds, i.e., birds chirping, rain, cars, people talking, and piano, onto the speech recordings.  Datasets with normal rate and without accent and background noise are denoted as SpeechRefer\_VF and SpeechNr3D\_VF (\textbf{V}ariations-\textbf{f}ree) accordingly. Our SpeechRefer dataset matches the size of ScanRefer and classifies scenes as ``unique" (containing a single object of the class) or ``multiple" (containing multiple objects). Similarly, the SpeechNr3D dataset mirrors Nr3D and is categorized as ``Easy" or ``Hard" based on the number of distractors, and as ``View-dependent" or ``View-independent" depending on the reliance on the speaker's viewpoint. The distribution of speech durations in the SpeechRefer dataset is illustrated in Figure \ref {fig:statistical}, showing a prominent peak in the 4–8 second range, which represents the most frequent duration span. Overall, durations vary from 1 to 35 seconds, though instances at the extremes—particularly those shorter than 3 seconds or longer than 15 seconds—are comparatively rare. The average duration across all samples is 5.89 seconds. To further evaluate the robustness of SpeechRefer, we manually recorded an additional set of speech data comprising 699 training samples and 352 validation samples, designed to reflect real-world usage scenarios more accurately. The corresponding evaluation results are presented in \ref{Experimental Results}. 

\textbf{Evaluation metrics.} We adopt the same metrics as other T-3DVG methods: Acc@0.25IoU and Acc@0.5IoU, which measure the percentage of correctly predicted bounding boxes with IoU scores exceeding 0.25 and 0.5. For SpeechNr3D, we evaluate the percentage of successful matches between predicted and ground truth bounding boxes, following previous work\cite{referit3d}.

\textbf{Implementation details.} We train the model end-to-end on a single NVIDIA RTX4090. The optimizer and learning rate are consistent with the T-3DVG methods integrated with SpeechRefer. For 3DVG-Transformer, we train for 200 epochs with a batch size of 2, pairing each scene with 32 sentences. For M3DRefCLIP, the network is trained for 60 epochs with a batch size of 4, pairing each scene with 8 sentences. Other training configurations align with standard T-3DVG methods.
\subsection{Experimental Results}
\label{Experimental Results}
Tabel \ref{tab:scanrefer results} presents the results of 3DVG-Transformer\cite{3dvgtransformer2021} and M3DRefCLIP\cite{Multi3DRefer} integrated with our SpeechRefer on the SpeechRefer\_CA dataset, alongside baseline results from four methods\cite{scanrefer,EDA,3dvgtransformer2021,Multi3DRefer}. 

The baseline performance suffers significantly when compared to results using precise text inputs due to errors in transcribed text. In contrast, SpeechRefer shows marked improvements. For 3DVG-Transformer, SpeechRefer achieves gains of 2.23\%/1.76\% in overall accuracy. 
For M3DRefCLIP, the overall accuracy improves by more than 2.30\%/1.60\%, with a notable 2.80\%/2.10\% gain on the ``multiple" subset, demonstrating SpeechRefer's ability to effectively distinguish distractors. 
Further fine-tuning on the variations-free SpeechRefer\_VF dataset yields even better results than training directly on it (please refer to the results in supplementary materials). This highlights SpeechRefer's robustness in handling variations.
Quantitative results demonstrate that our SpeechRefer can reliably localize target objects from speech descriptions and can easily integrate with existing T-3DVG methods, significantly enhancing performance and robustness.
For the SpeechNr3D\_CA dataset, as shown in Table \ref{tab:nr3d results}, our results outperform the baseline across all subsets, with an overall accuracy improvement of 3.60\%. Compared to the SpeechRefer dataset, SpeechNr3D is more sensitive to transcription errors due to its shorter descriptions (11.4 words on average, compared to ScanRefer's 20.27 words). The shorter descriptions make transcription errors more impactful, but SpeechRefer mitigates these adverse effects effectively, demonstrating robust performance. 

Furthermore, we evaluate SpeechRefer on the aforementioned real-world dataset, which closely mirrors practical application scenarios, with the results summarized in Table \ref{tab:real}. The experiments show that SpeechRefer can effectively adapt to real-world speech inputs with a relatively small amount of fine-tuning. Specifically, it's fine-tuned on 699 manually recorded training samples and evaluated on 352 validation samples. For comparison, the results in the first row of Table \ref{tab:real} are obtained using the same number of synthesized validation samples, which are drawn from the introduced SpeechRefer\_CA dataset, further demonstrating the model’s robustness and generalization ability in realistic settings.

\begin{table}
        \renewcommand\arraystretch{1}
	\centering
        \caption{Performance on real speech data. Both evaluations—on synthetic and human-recorded speech—were conducted using the same 352-sample validation set. The synthetic speech samples were taken directly from the SpeechRefer\_CA dataset, while the real-world speech consisted of manually recorded audio collected from human speakers.}
	{\begin{tabular}{ccc}	
            \noalign{\smallskip}\hline\noalign{\smallskip}
            \rowcolor{gray!20}
             Methods & Unique & Data\\
            \noalign{\smallskip}\hline\noalign{\smallskip}
            3DVG-T  &75.75/58.18 & Synthetic speech \\
		  +SpeechRefer (ours) &89.69/67.27 & Real-world speech \\     
  \noalign{\smallskip}\hline
	\end{tabular}}
	\label{tab:real}
\end{table}

\begin{figure*}
\centering
\includegraphics[width=\textwidth]{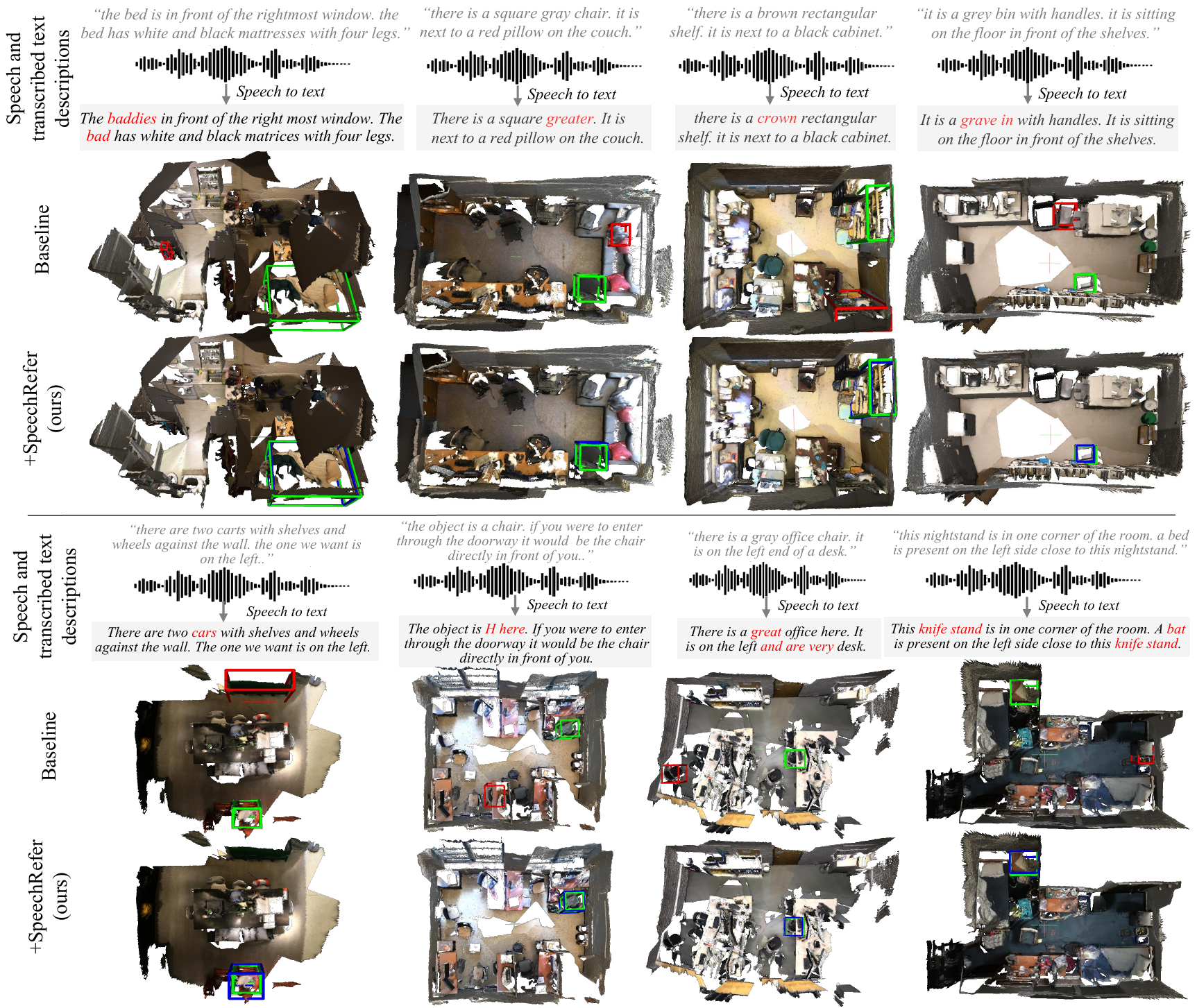}
\caption{Qualitative results. We utilize \textcolor{green}{green}, \textcolor{red}{red}, and \textcolor{blue}{blue} boxes to represent the ground truth, the prediction of baseline, and the prediction of SpeechRefer based on 3DVG-Transformer, respectively.}
\label{fig:visualization}
\end{figure*}
\subsection{Ablation Studies}
\textbf{Speech rate and background noise.}  We evaluate SpeechRefer's robustness under varied noisy conditions by introducing two additional variations: speech rate and background noise. For speech rate, we tested slower (100) and faster (200) rates. Results in Table \ref{tab:speechrate} show improved performance at slower rates, where reduced speed enhances semantic clarity. Even at faster rates, SpeechRefer maintains strong performance, showcasing its resilience. Under background noise conditions, SpeechRefer consistently outperforms baseline, which indicates its robustness in real-world noisy scenarios.
\begin{table}
        \renewcommand\arraystretch{1}
	\centering
    \caption{Ablation study of SpeechRefer's robustness under different conditions, including varying speech rate and background noise, based on 3DVG-Transformer.}
	{\begin{tabular}{ccccc}	
            \noalign{\smallskip}\hline\noalign{\smallskip}
            \rowcolor{gray!20}
            & & Unique & Multiple & Overall \\ 
            \rowcolor{gray!20}
             \multicolumn{2}{c}{\multirow{-2}*{Type of variations}} & 0.25 & 0.25 & 0.25\\
            \noalign{\smallskip}\hline\noalign{\smallskip}
             \multirow{2}{*}{Speech Rate}&100 & 78.44 & 40.43 & 47.81 \\
             &200 & 77.13 & 39.74 & 47.00\\
            \noalign{\smallskip}\hline\noalign{\smallskip}
            \multicolumn{2}{c}{Background Noise} & 77.48&40.50&47.67\\
		\noalign{\smallskip}\hline
	\end{tabular}}
    \label{tab:speechrate}
\end{table}
\textbf{Module contributions.} Table \ref{tab:ab each module} provides ablation results for different module combinations based on 3DVG-Transformer. The second row highlights the significant gains achieved by the speech learnable layers, which capture subtle differences in addition to acoustic similarities, thereby enhancing speech feature expressiveness. 
From the final row, we can see that the combination of the confidence-based complementary module and the contrastive complementary module further improves performance. This validates their roles in mitigating transcription errors and enhancing robustness.


\begin{table}
        \renewcommand\arraystretch{1}
	\centering
        \caption{Ablation study of different module combinations in SpeechRefer: SLL (speech learnable layers), CBM (confidence-based complementary module), CCM (contrastive complementary module). Evaluated on SpeechRefer\_CA datasets.}
	{\begin{tabular}{ccccccc}	
            \noalign{\smallskip}\hline\noalign{\smallskip}
            \rowcolor{gray!20}
             &  &   & \multicolumn{2}{c}{Multiple} & \multicolumn{2}{c}{Overall}\\
             \rowcolor{gray!20}
            \multirow{-2}*{SLL}&\multirow{-2}*{CBM}&\multirow{-2}*{CCM} & 0.25 & 0.5 & 0.25 & 0.5 \\
            \noalign{\smallskip}\hline\noalign{\smallskip}
     	 & & & 37.27&27.02 & 44.77&32.49\\
             \checkmark & & & 38.52&27.72 & 45.92&33.32 \\
             \checkmark & & \checkmark & 38.33&28.13 & 45.56&33.47 \\
             \checkmark &\checkmark & \checkmark & \textbf{39.74}&\textbf{28.73} & \textbf{47.00}&\textbf{34.25}\\
		\noalign{\smallskip}\hline
	\end{tabular}}
    \label{tab:ab each module}
\end{table}
\textbf{Alignment types in Contrastive Complementary Module.} Table \ref{tab:ab losstype} illustrates performance across alignment types (T: text features; O: visual features; S: speech features). When aligning only speech and visual features or only text and visual features, the performance is not optimal.
\begin{table}
        \renewcommand\arraystretch{1}
	\centering
        \caption{Performance cross alignment types (T represents text features, O represents visual features, S represents speech features). Rows 1–4: SpeechRefer + 3DVG-Transformer; Rows 5–6: SpeechRefer + M3DRefCLIP. }
	{\begin{tabular}{ccccccc}	
            \noalign{\smallskip}\hline\noalign{\smallskip}
             \rowcolor{gray!20}
            T\&O & S\&O & T\&S & Unique & Multiple & Overall \\
            \noalign{\smallskip}\hline\noalign{\smallskip}
     	 & & & 75.84/55.78 & 38.29/27.63 & 45.57/33.09 \\
             \checkmark & & & 75.64/54.51 & 36.74/25.98 & 44.29/31.51 \\
             \checkmark & & \checkmark & 76.17/54.98 & 36.33/26.37 &44.06/31.92 \\
             \checkmark &\checkmark & \checkmark & 77.13/57
             19& 39.74/28.73 & 47.00/34.25 \\
             \noalign{\smallskip}\hline\noalign{\smallskip}
             \checkmark & & &84.60/74.50 & 42.00/35.20 & 50.30/42.80 \\
             \checkmark & \checkmark & \checkmark & 85.20/74.10 & 44.80/37.30 & 52.60/44.40 \\
		\noalign{\smallskip}\hline
	\end{tabular}}
    \label{tab:ab losstype}
\end{table}

\textbf{Only Speech.} While transcribed text improves overall accuracy, it is not an ideal solution for the speech-guided 3D visual grounding task, as incorporating text adds complexity to the network and training process. Therefore, we conducted a preliminary exploration of a simplified SpeechRefer architecture that relies solely on the speech and point cloud modality. The results, presented in Table \ref{tab:only speech}, show that our approach significantly outperforms AP-Refer\cite{aprefer}, the first and only 3DVG network relies solely on the speech and point cloud. This demonstrates the effectiveness and robustness of our SpeechRefer framework.

We also explore the impact of different speech encoders. Table \ref{tab:4} presents the results of the speech-only network using Imagebind and Whisper as speech encoders, respectively. The results indicate that when Imagebind is used as the encoder, the performance is notably lower compared to Whisper. We attribute this discrepancy to the fact that, while Imagebind is designed to learn joint embeddings across six modalities, it primarily integrates these modalities through image-paired data. As a result, when applied to point cloud-centric tasks, a gap remains between the modalities. In contrast, Whisper is an encoder specifically optimized for speech processing, enabling it to more effectively capture the phonetic similarities within speech and guide the network toward better results.
\begin{table}
        \renewcommand\arraystretch{1}
	\centering
        \caption{Comparison results of AP-Refer\cite{aprefer} and simplified SpeechRefer integrate with 3DVG-Transformer (the second row) and M3DRefCLIP (the final row).}
	{\begin{tabular}{cccc}	
            \noalign{\smallskip}\hline\noalign{\smallskip}
            \rowcolor{gray!20}
            Methods & Unique & Multiple & Overall \\          
            \noalign{\smallskip}\hline\noalign{\smallskip}
            AP-Refer & 48.62/29.59 & 16.94/9.96 & 23.09/13.77 \\
            \noalign{\smallskip}\hline\noalign{\smallskip}
            \multirow{2}{*}{Ours} & 76.04/54.44 & 35.93/25.62 & 43.72/31.22 \\
             & 84.00/73.90 & 41.60/34.50 & 49.50/42.20 \\
		\noalign{\smallskip}\hline
	\end{tabular}}
	\label{tab:only speech}
\end{table}
\begin{table}
        \renewcommand\arraystretch{1}
	\centering
        \caption{Ablation study of speech encoder. Results on different speech encoders (Imagebind\cite{imagebind} and Whisper\cite{whisper}), evaluating on 3DVG-Transformer.}
	{\begin{tabular}{cccc}	
            \noalign{\smallskip}\hline\noalign{\smallskip}
            \rowcolor{gray!20}
             Encoder & Unique & Multiple & Overall\\
            \noalign{\smallskip}\hline\noalign{\smallskip}
		  Imagebind & 55.10/39.98 & 22.63/16.38 & 28.93/20.96 \\
            Whisper & 76.04/54.44 & 35.93/25.62 & 43.72/31.22 \\      
  \noalign{\smallskip}\hline
	\end{tabular}}
	\label{tab:4}
\end{table}
\textbf{Ablation study of confidence score weight.} Table \ref{tab:3} presents the results from various weight combinations assigned to the speech confidence score $S_a$ and the text confidence score $S_t$ in computing the final confidence output $S$. Among the configurations tested, the best performance was achieved with equal weights of 0.5 for both modalities. In contrast, relying solely on either the text or speech confidence score resulted in significantly lower performance. This highlights the importance of both speech and text information in guiding the network, demonstrating that a balanced integration of these modalities yields the most effective results. 
\begin{table}
        \renewcommand\arraystretch{1.2}
	\centering
        \caption{Ablation study on confidence score weighting with various combinations of the speech score $S_s$ and transcribed text score $S_t$. Five different weight configurations were tested. Evaluated on SpeechRefer\_CA dataset.}
	{\begin{tabular}{ccccc}	
            \noalign{\smallskip}\hline\noalign{\smallskip}
            \rowcolor{gray!20}
            $S_s$ & $S_t$  & Unique & Multiple & Overall\\
            \noalign{\smallskip}\hline\noalign{\smallskip}
		  0.0 & 1.0 & 75.84/55.78 & 38.29/27.63 & 45.57/33.09 \\
            0.2 & 0.8 & 77.01/56.75 & 39.52/28.61 & 46.80/34.07 \\
		0.5 & 0.5 & \textbf{77.13/57.19} & \textbf{39.74/28.73} & \textbf{47.00/34.25} \\    
  		0.8 & 0.2 & 75.32/55.96 & 38.28/27.68 & 45.46/33.17\\    
		1.0 & 0.0 & 73.15/54.66 & 36.70/26.60 & 43.78/32.05 \\    
            \noalign{\smallskip}\hline
	\end{tabular}}
	\label{tab:3}
\end{table}
\subsection{Visualization and Limitations}
To further demonstrate the effectiveness of SpeechRefer, Figure \ref{fig:visualization} illustrates qualitative comparisons between our method and the baseline on 3DVG-Transformer.  
Transcription errors often lead the baseline to misidentify objects, while SpeechRefer overcomes these challenges to successfully identify the target objects.

\begin{figure}
    \centering
    \includegraphics[width=\linewidth]{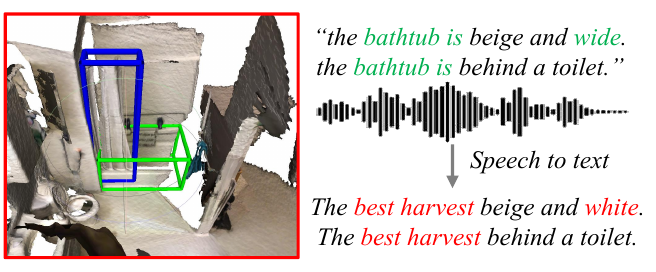}
    \caption{Visualization of a failure case in the SpeechRefer\_CA dataset. The \textcolor{green}{green} boxes and \textcolor{blue}{blue} boxes represent the ground truth object and the wrong prediction of our SpeechRefer based on 3DVG-Transformer, respectively.}
    \label{fig:limitations}
\end{figure}

Despite the promising performance, our method still has limitations. As shown in Figure \ref{fig:limitations}, significant transcription errors, such as consecutive mistranscribed words caused by accents or background noise, can cause the textual features to deviate too much, leading the model to misidentify targets. 
While text features improve overall accuracy, reliance on text is not ideal for speech-guided 3D visual grounding tasks. Future work will explore improved extraction and utilization of semantic information directly from speech, enabling reliable reasoning without text dependency.

\section{Conclusion}
\label{sec:conclusion}
We propose SpeechRefer, a novel 3D visual grounding framework that bridges the gap between speech and 3DVG in practical noisy environments. By leveraging speech features as a complement to transcribed text features derived from speech input, our method addresses the challenges of imperfect speech-to-text transcriptions, enhancing both performance and robustness. 
Our method integrates seamlessly with existing T-3DVG methods, as demonstrated through extensive experiments on our SpeechRefer and SpeechNr3D datasets, where it consistently outperforms baselines by a large margin. 
More importantly, our work provides additional insights distinct from previous 3D visual grounding methods—ambiguous speech in practical noisy environments introduces more possibilities than text, which enables the system to better infer user intent amidst a variety of possibilities and choices, rather than being constrained by definitive text.
Overall, as the first attempt to localize a target object using noisy and ambiguous speech input, SpeechRefer shows promising results with room for improvement, paving the way for more intuitive 3D visual grounding.
Future work will focus on deeper speech learning and better speech-visual integration to robust performance without text reliance.

\bibliographystyle{IEEEtran}
\bibliography{reference.bib}

\end{document}